\documentclass[11pt]{article}

\usepackage[final]{acl}

\usepackage{times}
\usepackage{latexsym}

\usepackage[T1]{fontenc}

\usepackage[utf8]{inputenc}

\usepackage{microtype}

\usepackage{inconsolata}

\usepackage{graphicx}

\title{DWTSumm: Discrete Wavelet Transform for Document Summarization}

\author{%
Rana Aref Salama\textsuperscript{1,2}, Abdou Youssef\textsuperscript{1}, and  Mona Diab\textsuperscript{3}\\
{\normalsize \textsuperscript{1} School of Engineering and Applied Science, George Washington University, USA}\\
{\normalsize \textsuperscript{2} Faculty of Computers and Artificial Intelligence, Cairo University, Egypt}\\
{ \normalsize \textsuperscript{3} Language Technologies Institute, Carnegie Mellon University, USA}
}

\begin{document}
\maketitle
\begin{abstract}
Summarizing long, domain-specific documents with large language models (LLMs) remains challenging due to context limitations, information loss, and hallucinations, particularly in clinical and legal settings. We propose a Discrete Wavelet Transform (DWT)-based multiresolution framework that treats text as a semantic signal and decomposes it into global (approximation) and local (detail) components. Applied to sentence- or word-level embeddings, DWT yields compact representations that preserve overall structure and critical domain-specific details, which are used directly as summaries or to guide LLM generation. Experiments on clinical and legal benchmarks demonstrate comparable ROUGE-L scores. Compared to a GPT-4o baseline, the DWT based summarization consistently improve semantic similarity and grounding, achieving gains of over 2\% in BERTScore, more than 4\% in Semantic Fidelity, factual consistency in legal tasks, and large METEOR improvements indicative of preserved domain-specific semantics. Across multiple embedding models, Fidelity reaches up to 97\%, suggesting that DWT acts as a semantic denoising mechanism that reduces hallucinations and strengthens factual grounding. Overall, DWT provides a lightweight, generalizable method for reliable long-document and domain-specific summarization with LLMs.
\end{abstract}

\section{Introduction}

Large Language Models (LLMs) have become the dominant paradigm for text summarization \citep{raffel2020t5,brown2020language}. However, summarizing complex documents remains challenging when inputs are either long or highly domain-specific, such as clinical records or legal opinions, where factual accuracy and contextual grounding are critical. These challenges are further amplified when documents exceed the model’s limited context window, forcing text truncation, aggressive compression, or multi-stage summarization pipelines 
\citep{beltagy2020longformer,dao2022flashattention}.
\\
Hierarchical summarization, which segments documents and recursively merges intermediate summaries, is a common strategy for handling long inputs \citep{wu2021,chang2024}. While effective for length reduction, this approach often introduces hallucinations and information loss, particularly in domain-specific settings \citep{maynez2020faithfulness,ji2023survey}. These limitations highlight the need for methods that preserve semantic structure and factual information across diverse document lengths and domains.
\\

In this paper, we propose a summarization framework based on the \emph{Discrete Wavelet Transform (DWT)}. DWT is a well-established signal processing technique \citep{mallat1989theory,daubechies1992ten} that recently demonstrated its effectiveness in capturing semantic structure in natural language representations \citep{my}. We apply DWT to sentence-level representations to derive compact document encodings that preserve essential semantic information while reducing redundancy.
\\

We argue that DWT is particularly well suited for summarization for three reasons. First, its \emph{multiresolution analysis} provides hierarchical semantic representations without requiring recursive LLM inference. Second, its \emph{compression capability} enables substantial sequence length reduction—often exceeding 50--75\%—while preserving semantic fidelity. Third, its \emph{structural separation of global and local information} supports factual grounding in both long-document and domain-specific summarization tasks.
\\
\\

To the best of our knowledge, this work is the first to systematically introduce DWT-based compression for LLM-based summarization. Our experiments demonstrate that DWT improves semantic consolidation and factual consistency while enabling efficient summarization of long and domain-specific documents.

\paragraph{Contributions.}
\begin{enumerate}
\item We introduce a DWT-based multi-resolution compression framework for summarizing long legal documents and clinical documents with LLMs.
\item We show that DWT preserves global context and domain-specific details in these domains while substantially reducing input length.
\item We demonstrate that integrating DWT with LLMs improves semantic fidelity and reduces hallucinations in both legal and clinical summarization tasks.

\end{enumerate}

\section{Motivation}
The Discrete Wavelet Transform (DWT) provides a multiresolution representation that separates global document structure from localized, domain-specific details. This property is particularly relevant for legal and clinical documents, which are often long, hierarchically organized, and sensitive to factual inaccuracies. By explicitly modeling both global context and local information, DWT may help alleviate challenges associated with long-context processing, including the lost-in-the-middle problem~\citep{lost}. Rather than relying solely on long-range attention over flat sequences, DWT restructures long documents into compact representations that preserve both overarching context and salient local details. Additionally, by isolating fine-grained, domain-specific information, DWT can help reduce the tendency of LLMs to fill informational gaps with unsupported or hallucinated content, thereby strengthening factual grounding in the generated summaries.
\\
\section{Related Work}

\subsection{Long Text Summarization}
Recent work on long-document summarization has shifted toward structured and evidence-aware pipelines aimed at improving faithfulness and reducing hallucinations. These include evidence-guided merging, token- and chunk-level compression methods such as LLMLingua ~\citep{llmlingua} , and hybrid long-context and retrieval strategies that address the lost-in-the-middle problem \citep{lost,li2024hybrid}. Collectively, these advances emphasize designing summarization pipelines that explicitly preserve structure, evidence, and reliability rather than relying solely on longer context windows.
\\
\subsection{Domain-Specific Summarization}

\paragraph{Clinical Summarization.}
Clinical summarization has become essential for managing growing volumes of patient data. Recent approaches leverage large pretrained language models, retrieval-augmented generation, and medical knowledge graphs to produce concise and evidence-grounded summaries \citep{cs1,longhealth,vveen2024,MedGraphRAG}. To handle long and longitudinal records, hierarchical and multiresolution models combined with retrieval-based filtering have been proposed \citep{lopez2025,clinicsum}. Despite these advances, ensuring factual accuracy and clinical fidelity in summaries of complex records remains an open challenge \citep{factpico,tang2024}.
\\
\paragraph{Legal Summarization.}
In the legal domain, summarization is increasingly important for managing large and complex legal texts \citep{deroy2023}. Recent work moves beyond extractive methods toward domain-specific generative models that incorporate discourse structure and legal reasoning \citep{shukla2024}. To handle long and hierarchical narratives, multiresolution modeling and Legal Retrieval-Augmented Generation (Legal-RAG) have been adopted to mitigate the \emph{lost-in-the-middle} problem and reduce hallucinations \citep{bhattacharya2025,deroy2025}.
\\

\section{Discrete Wavelet Transform}

DWT is a multiresolution analysis framework that hierarchically decomposes a signal $f(t)$ through recursive filtering and downsampling. At each level $j$, $f(t)$ is convolved with low-pass and high-pass filters to produce \textit{Approximation coefficients} ($cA_j$), representing low varying components, and \textit{Detail coefficients} ($cD_j$), capturing fine-scale or high variations \citep{69,73}. This decomposition is governed by a \textit{Mother Wavelet} (MW) $\psi(t)$, scaled by $a$ and translated by $b$ \cite{69}:

\begin{equation}
\psi_{a,b}(t) = \frac{1}{\sqrt{a}} \, \psi\Big(\frac{t-b}{a}\Big).
\end{equation}

Where \(a\) controls the scale (resolution) and \(b\) the position along the signal. The wavelet coefficients are computed by projecting the input signal onto a MW, such that the approximation and detail components \(cA_j(t)\) and \(cD_j(t)\) are derived from how well the wavelets match the local patterns in \(f(t)\) at different scales. Smaller scales \(a\) capture fine, high-frequency details, while larger scales capture coarse, low-frequency structures.
\\

Conceptually, DWT is similar to convolutional neural networks (CNNs) in using sliding-window filters and downsampling \citep{74}, but unlike CNNs, the filters in DWT are analytically defined rather than learned, providing interpretability and stability.
\\

However, DWT filters are analytically defined rather than learned, ensuring deterministic interpretability. 
While various wavelet families exist (e.g., Haar, Symmlets, Coiflets) \citep{20.1191457}, this study utilizes the Daubechies (db) wavelet to demonstrate the feasibility of the proposed DWT-based summarization framework without the confounding variables of wavelet optimization.
\footnote{For a more detailed explanation of Wavelet Transform theory, refer to~\citep{73,20.1191457,68.brunton_kutz_2019}.}

\section{DWT-Text Summarization}

We propose a multi-resolution hybrid summarization framework that models text as a discrete semantic signal \(X\), where sentence-level embeddings form a sequential vector representation. The Discrete Wavelet Transform (DWT) is applied to hierarchically decompose this signal into approximation coefficients (\(cA\)), which capture the global semantic structure and narrative flow of the document, and detail coefficients (\(cD\)), which capture localized, domain-specific information and salient semantic variations.
\\
The framework follows four stages. First, the input document is embedded using a domain-specific encoder to obtain a continuous semantic representation. Second, a multi-level DWT is applied to produce a hierarchy of coefficients \(\{cA_L, cD_L, \dots, cD_1\}\), where the decomposition level is selected based on document length and the desired compression ratio. Third, these coefficients are mapped back to representative sentences through nearest-neighbor retrieval in the embedding space, yielding a compact semantic skeleton from \(cA\) and a set of salient detail sentences from \(cD\). Finally, the resulting multi-resolution representation is either used directly as a compressed summary that is refined by a large language model (LLM), or incorporated into the LLM prompt to guide abstractive generation, explicitly grounding the model in both global context and critical details.
\\

This design positions DWT as a semantic compression and structuring mechanism rather than a generative model, allowing the LLM to focus on fluent synthesis while operating over a factually grounded and information-dense representation. Consequently, the framework enables effective summarization of long and domain-specific documents while mitigating information loss and hallucinations.

\begin{figure*}
    \centering
    \includegraphics[width=0.99\linewidth]{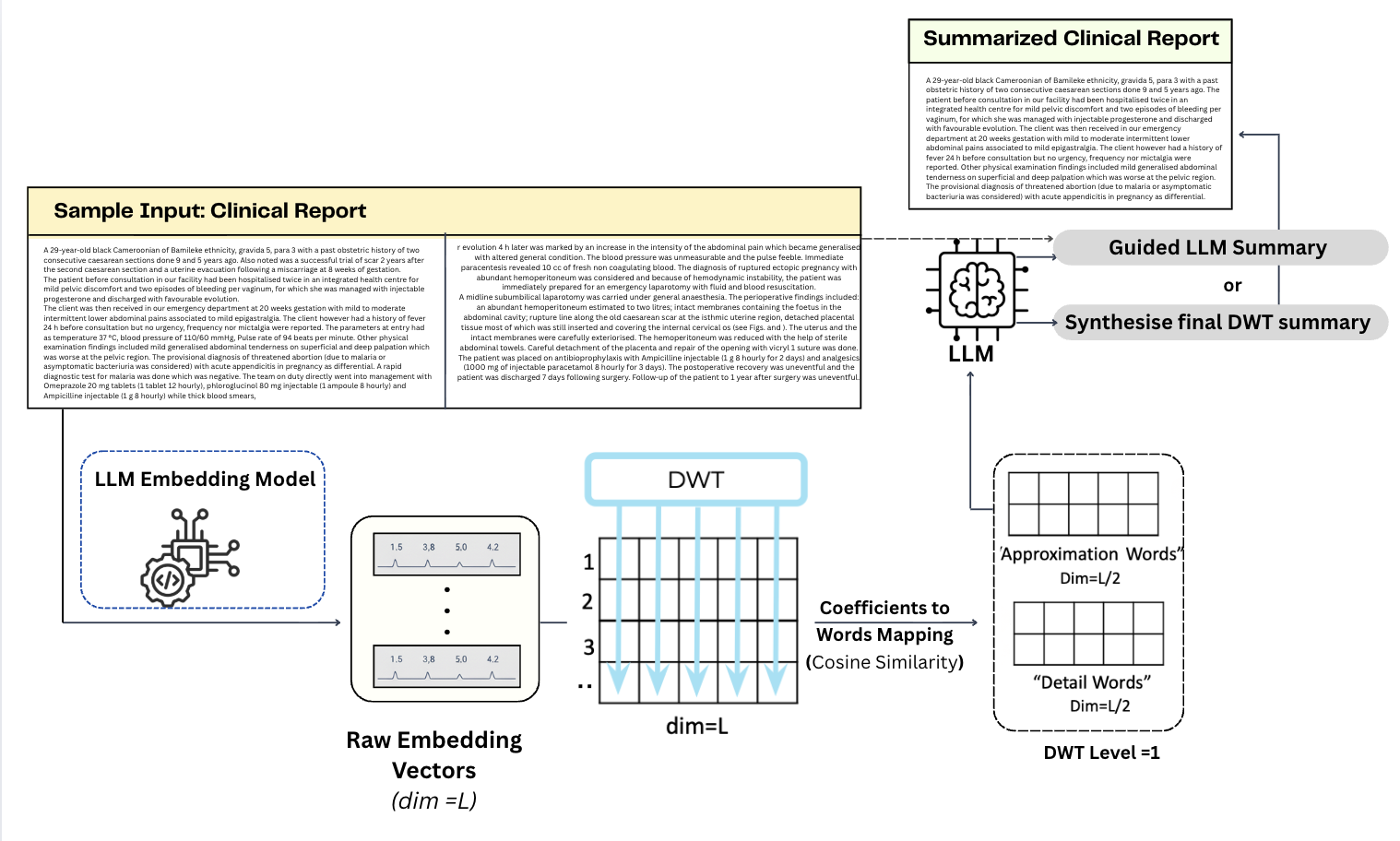}
    \caption{Overview of the DWT-based Summarization Framework. Long-context documents are decomposed via DWT into approximation ($cA$) and detail ($cD$) components, capturing global semantic structure and localized information variations to support more effective LLM-based summarization.}
    \label{fig:placeholder}
\end{figure*}

\section{Evaluation}
To evaluate the effectiveness of the proposed model, we conduct experiments across clinical summarization and legal summarization.

\subsection{Clinical Summarization}

\paragraph{Experimental Setting}

To evaluate the proposed framework on clinical summarization, we conduct experiments on the MultiClinSum dataset \citep{multiclinsum}, a large-scale multilingual corpus of clinical case reports designed to benchmark the preservation of critical medical information in automated summaries. Sentence- and word-level representations are obtained using BioClinical ModernBERT embeddings\footnote{\texttt{NeuML/bioclinical-modernbert-base-embeddings}}. We use GPT-4o ~\citep{gpt4o} as the base model.
We evaluate generated summaries using a combination of lexical, semantic, and factual metrics. Lexical overlap is measured using average ROUGE-L \citep{rouge} and METEOR \citep{meteor}, while semantic similarity is assessed using average BERTScore \citep{bertscore}. Factual consistency is evaluated using the G-Eval framework \citep{geval}. In addition, we measure semantic fidelity by computing the cosine similarity between the embedding of the original document and that of the generated summary.

Together, these metrics capture both linguistic quality and semantic preservation, with the fidelity measure specifically assessing the ability of the DWT-based approach to retain salient information that is often lost in long-context truncation.

\begin{table*}[t]
\centering
\small
\setlength{\tabcolsep}{11pt}
\caption{Evaluation results on the MultiClinSum dataset. We report performance of the proposed DWT-based summarization framework using either DWT alone or a hybrid DWT+GPT model for summary generation, compared against a GPT-4o baseline as reported in \cite{multiclinsum}. All metrics are reported as percentages. Bold values denote improvements over the baseline, and red indicates the best overall performance across all configurations.}
\label{tab:results}
\begin{tabular}{lcccccc}
\hline
\textbf{Model} & \textbf{Comp. Ratio} & \textbf{ROUGE-L} & \textbf{METEOR} & \textbf{BERTScore} & \textbf{Fidelity} & \textbf{Fact.} \\ \hline
GPT-4o (Baseline) & -- & 30.77 & 32.60 & 85.46 & 91.60 & 4.0 \\
DWT (Only) & 69.41 & 30.08 & \textbf{33.30} & \textbf{86.17} & \textbf{94.38} & \textbf{5.0} \\
DWT + GPT & 61.30 & \textcolor{red}{\textbf{30.79}} & \textcolor{red}{\textbf{46.00}} & \textcolor{red}{\textbf{87.89}} & \textcolor{red}{\textbf{95.06}} & \textbf{5.0} \\ \hline
\end{tabular}
\end{table*}

\paragraph{Experimental Results}

To evaluate the proposed approach, we consider two configurations: using DWT-generated summaries independently and using them to guide GPT-based summarization. Both settings are compared against a strong baseline consisting of GPT-4o without DWT augmentation. As shown in Table~\ref{tab:results}, the DWT-only model produces competitive summaries while achieving a compression ratio exceeding 60\%, outperforming the baseline in BERTScore, Fidelity, factual consistency, and METEOR.

These results indicate that DWT effectively captures both global structure and fine-grained, domain-specific semantics in long clinical documents, enabling substantial compression without loss of critical information.

The benefits of DWT are further amplified in the hybrid DWT+GPT configuration, which consistently improves over the baseline across semantic and factual metrics, yielding gains of over 2\% in BERTScore and more than 4\% in Fidelity while maintaining comparable ROUGE-L performance and a compression ratio above 60\%. The improvement in METEOR further suggests that DWT preserves nuanced clinical semantics, providing salient and fact-grounded representations that enhance downstream LLM-based summarization.

Table~\ref{tab:clinical_example} illustrates an example of a long clinical case description and its compressed representation, highlighting how salient clinical facts are preserved while redundant narrative content is removed.  The compressed summary preserves key clinical facts, including patient history, diagnosis, intervention, and outcome, while removing redundant narrative content, illustrating the ability of DWT-based semantic compression to retain clinically salient information under substantial length reduction.

\begin{table*}[ht]
\centering
\tiny
\caption{Example of long clinical text and its DWT based summarization using DWT+GPT model.}
\label{tab:clinical_example}
\begin{tabular}{p{0.47\textwidth} p{0.47\textwidth}}
\hline
\textbf{Original Clinical Text (Excerpt)} & \textbf{Compressed Clinical Summary} \\
\hline
A 29-year-old black Cameroonian of Bamileke ethnicity, gravida 5, para 3 with a past obstetric history of two consecutive caesarean sections done 9 and 5 years ago. Also noted was a successful trial of scar 2 years after the second caesarean section and a uterine evacuation following a miscarriage at 8 weeks of gestation.
The patient before consultation in our facility had been hospitalised twice in an integrated health centre for mild pelvic discomfort and two episodes of bleeding per vaginum, for which she was managed with injectable progesterone and discharged with favourable evolution.
The client was then received in our emergency department at 20 weeks gestation with mild to moderate intermittent lower abdominal pains associated to mild epigastralgia. The client however had a history of fever 24 h before consultation but no urgency, frequency nor mictalgia were reported. The parameters at entry had as temperature 37 °C, blood pressure of 110/60 mmHg, Pulse rate of 94 beats per minute. Other physical examination findings included mild generalised abdominal tenderness on superficial and deep palpation which was worse at the pelvic region. The provisional diagnosis of threatened abortion (due to malaria or asymptomatic bacteriuria was considered) with acute appendicitis in pregnancy as differential. A rapid diagnostic test for malaria was done which was negative. The team on duty directly went into management with Omeprazole 20 mg tablets (1 tablet 12 hourly), phloroglucinol 80 mg injectable (1 ampoule 8 hourly) and Ampicilline injectable (1 g 8 hourly) while thick blood smears, urinalysis, obstetric ultrasound, and full blood count were requested for the next morning.
Her evolution 4 h later was marked by an increase in the intensity of the abdominal pain which became generalised with altered general condition. The blood pressure was unmeasurable and the pulse feeble. Immediate paracentesis revealed 10 cc of fresh non coagulating blood. The diagnosis of ruptured ectopic pregnancy with abundant hemoperitoneum was considered and because of hemodynamic instability, the patient was immediately prepared for an emergency laparotomy with fluid and blood resuscitation.
A midline subumbilical laparotomy was carried under general anaesthesia. The perioperative findings included: an abundant hemoperitoneum estimated to two litres; intact membranes containing the foetus in the abdominal cavity; rupture line along the old caesarean scar at the isthmic uterine region, detached placental tissue most of which was still inserted and covering the internal cervical os (see Figs. and ). The uterus and the intact membranes were carefully exteriorised. The hemoperitoneum was reduced with the help of sterile abdominal towels. Careful detachment of the placenta and repair of the opening with vicryl 1 suture was done. The patient was placed on antibioprophylaxis with Ampicilline injectable (1 g 8 hourly for 2 days) and analgesics (1000 mg of injectable paracetamol 8 hourly for 3 days). The postoperative recovery was uneventful and the patient was discharged 7 days following surgery. Follow-up of the patient to 1 year after surgery was uneventful.

 &
29-year-old gravida 5 para 3 at 20 weeks gestation with prior two caesarean sections presented with abdominal pain and clinical deterioration. Emergency laparotomy revealed uterine rupture at previous caesarean scar with massive hemoperitoneum and intact membranes. Surgical repair and antibiotic therapy were performed with uneventful recovery. \\
\hline
\end{tabular}
\end{table*}

\subsection{Legal Summarization}
To evaluate the proposed DWT-based summarization framework in the legal domain, we conduct experiments on the CaseSumm dataset \citep{heddaya}, a large-scale corpus of 25.6K U.S. Supreme Court opinions paired with their official syllabuses. The documents average 4,983 tokens, with many exceeding standard Transformer context limits, providing a rigorous testbed for assessing DWT-based summarization on long legal texts.
\\
\paragraph{Experimental Setting}
We employ a domain-specific long-context sentence embedding model\footnote{\texttt{joelito/legal-english-longformer-base}}, pretrained on large-scale legal corpora, to encode lengthy judicial opinions. Final summaries are generated using \texttt{GPT-4o}, either by directly synthesizing DWT-generated representations or by using the DWT decomposition to guide the generation process. Model outputs are evaluated against gold-standard syllabuses using BERTScore, Meteor, Factual Consistency and Semantic fidelity metric to assess semantic alignment and source grounding. As a baseline, we adopt a zero-shot GPT-4o configuration as reported in \citep{heddaya}.
\\
\paragraph{Experimental Results}

\begin{table*}[ht]
\centering
\caption{Average evaluation results on the legal summarization task. We compare the zero-shot GPT-4o baseline with the proposed DWT-only and DWT+GPT frameworks.}
\label{tab:legal_results}
\begin{tabular}{lcccc}
\hline
\textbf{Model} & \textbf{METEOR} & \textbf{BERTScore} & \textbf{Fact. (1--5)} & \textbf{Fidelity} \\
\hline
Baseline (GPT-4o) & 30.80 & 87.00 & 4.40 & 92.80 \\
DWT (Only)        & \textbf{35.90} & \textbf{88.00} & \textbf{5.00} & \textbf{97.10} \\
DWT + GPT         & \textcolor{red}{\textbf{45.9}} &  \textcolor{red}{\textbf{89.00}} & \textbf{5.00} & \textbf{98.11} \\
\hline
\end{tabular}
\end{table*}

The results in Table~\ref{tab:legal_results} demonstrate the effectiveness of the proposed DWT-based framework for long legal document summarization. The DWT-only configuration already outperforms the zero-shot GPT-4o baseline across semantic and factual metrics, indicating that multiresolution decomposition alone can preserve both global structure and critical legal details under substantial compression. The hybrid DWT+GPT model further amplifies these gains, achieving the highest METEOR, BERTScore, and Fidelity scores while maintaining perfect factual consistency. Notably, the substantial improvement in METEOR suggests that DWT effectively captures domain-specific semantic nuances in legal text, confirming its ability to preserve fine-grained legal meaning alongside overall coherence when summarizing long documents.

\section{Discussion}

\subsection{Multi-level DWT Summarization}
A key advantage of the Discrete Wavelet Transform (DWT) is its multiresolution analysis capability, whereby the transformation can be applied recursively to achieve progressively higher compression while preserving the coarse semantic structure of the input. Each additional level of decomposition retains the main approximation components while discarding increasingly fine-grained details, resulting in more aggressive compression. 
\\

As illustrated in Figure~\ref{fig:placeholder}, we examine the impact of three DWT levels on the DWT+GPT model using BERTScore, Fidelity, and METEOR. Notably, the Level-3 DWT summary achieves an effective compression of 87.5\%. Despite this substantial reduction, semantic similarity remains stable across levels, while factual consistency exhibits a slight improvement at deeper decompositions. These results indicate that increased multiresolution decomposition preserves core semantic content while strengthening factual grounding, demonstrating the robustness of DWT for summarizing long and complex clinical documents.

\begin{figure}
    \centering
    \includegraphics[width=0.99\linewidth]{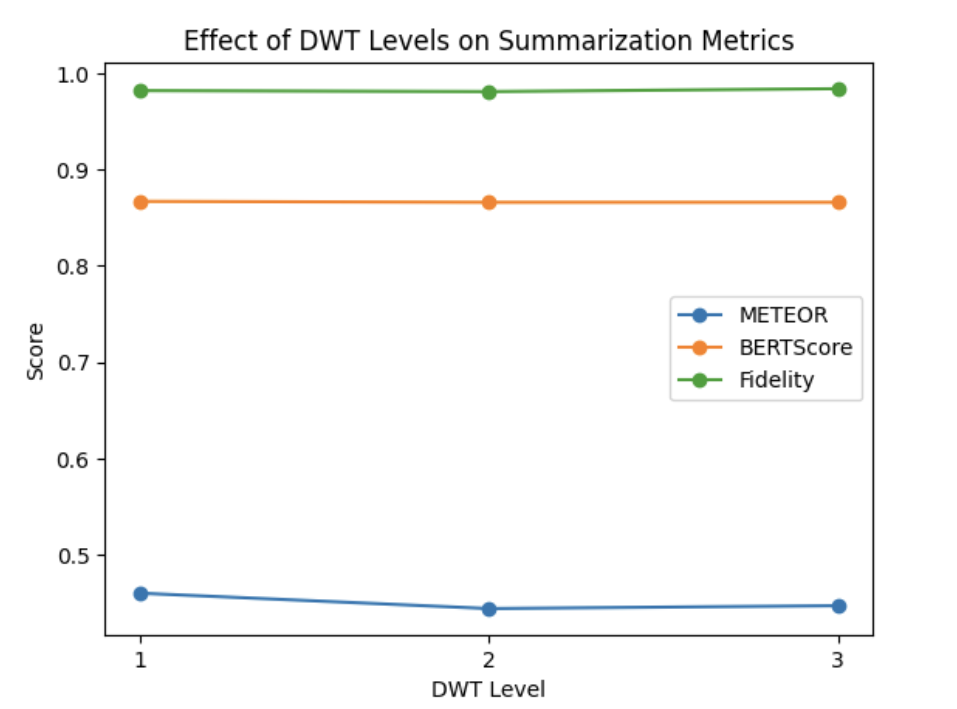}
    \caption{Effect of DWT decomposition level on summarization quality using MultiClinSum dataset. Results show stable semantic similarity and improved factual consistency at deeper decomposition levels, indicating effective multiresolution semantic preservation.}
    \label{fig:placeholder}
\end{figure}

\subsection{Efficacy of DWT Across Embeddings}
To assess the robustness and generalization capability of the proposed DWT-based framework, we examined its performance across multiple clinically oriented embedding models. As shown in figure by the results, DWT consistently preserves semantic and factual information regardless of the underlying embedding representation. Across embeddings—including MedCPT ~\citep{medcpt}, BioClinical-ModernBERT ~\citep{bioclinical}, BioLinkBERT , and BioClinical-ModernBERT-base ~\citep{biolinkbert} we observe stable BERTScore values and uniformly high Fidelity scores, all exceeding the baseline GPT performance. 
\\
Notably, Fidelity improves substantially over the baseline across all embeddings, reaching up to 98.66\% with BioClinical-ModernBERT-base, while BERTScore remains consistently strong, indicating preserved semantic similarity. This stability across heterogeneous embedding spaces suggests that DWT effectively captures core semantic structure in clinical text rather than overfitting to a specific representation. Overall, these findings support the view that DWT functions as a representation-agnostic multi-resolution mechanism, capable of extracting semantically meaningful and fact-preserving signals across diverse clinical contexts, thereby enhancing the generalization capacity of downstream summarization models.
\begin{figure}
    \centering
    \includegraphics[width=0.99\linewidth]{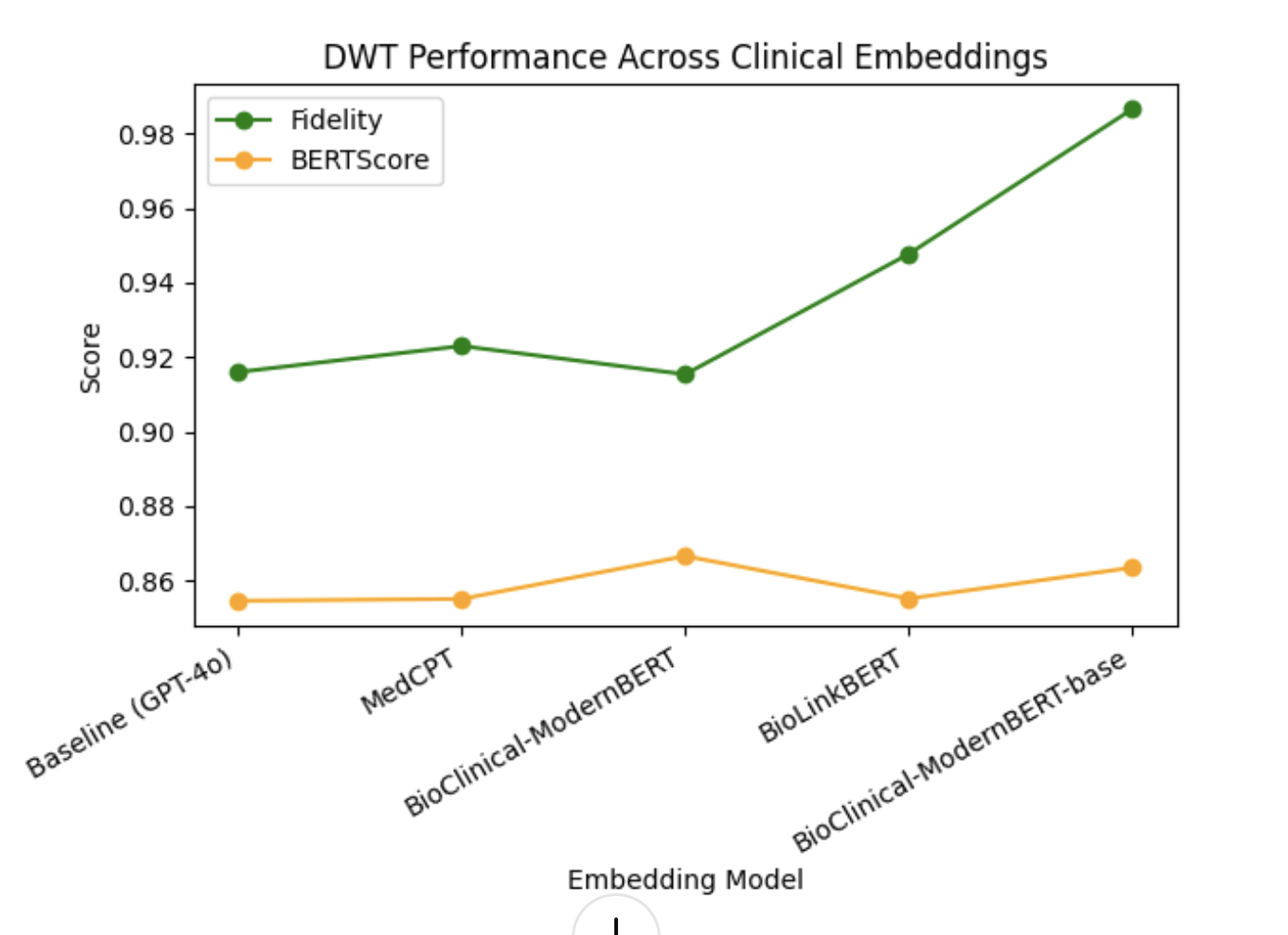}
    \caption{Performance of the DWT-based framework across clinical embedding models, showing consistent semantic preservation and improved factual grounding.}
    \label{fig:placeholder}
\end{figure}
\section{Conclusion}
In this work, we introduced a DWT-based  summarization framework for long and domain-specific documents, demonstrating its effectiveness across clinical and legal summarization tasks. Across all experiments, ROUGE-L scores remain consistent and comparable to strong LLM baselines, indicating that the proposed approach does not sacrifice surface-level overlap while operating under substantial compression. Importantly, the observed improvements in semantic and factual metrics suggest that DWT prioritizes the preservation of meaning rather than syntactic similarity, a behavior that is consistent with prior findings on wavelet-based semantic modeling \citep{my}.

Our results further show that DWT is not merely compressing text, but effectively \emph{denoising} long documents by filtering redundant or low-salience content while preserving semantically and factually critical information. This property leads to substantial gains in semantic fidelity and factual consistency, with fidelity scores reaching up to 97\% when paired with domain-specific embedding models. These findings indicate that DWT provides strong factual anchoring, reducing the tendency of LLMs to hallucinate when summarizing long and complex inputs.

We also evaluated the robustness of the framework across multiple domain-specific embedding models. The consistent performance across embeddings highlights the generalization capability of the DWT-based representation, suggesting that the approach captures semantic structure in a model-agnostic manner and can be readily adapted to different domains and representation spaces.

Overall, the proposed framework demonstrates that static semantic compression via DWT can significantly reduce input length while retaining multi-level semantic information, enabling efficient and reliable long-document summarization. By decoupling semantic structuring from language generation, DWT offers a lightweight, interpretable, and generalizable preprocessing mechanism that complements LLMs and can reduce computational cost while improving factual grounding in domain-specific summarization tasks.

\section*{Limitations}

\section{Limitations}

The proposed DWT-based framework has several limitations. First, its effectiveness depends on the quality of the underlying sentence or word embeddings; weaker representations may reduce the ability of DWT to capture salient semantic structure. Second, we adopt a fixed wavelet family and decomposition strategy for consistency, which may not be optimal across all domains or document types. Third, mapping wavelet coefficients back to text via nearest-neighbor retrieval introduces an approximation that may affect discourse coherence, although this is partially mitigated by downstream LLM synthesis.

\bibliography{custom}

@misc{longhealth,
      title={LongHealth: A Question Answering Benchmark with Long Clinical Documents}, 
      author={Lisa Adams and Felix Busch and Tianyu Han and Jean-Baptiste Excoffier and Matthieu Ortala and Alexander Löser and Hugo JWL. Aerts and Jakob Nikolas Kather and Daniel Truhn and Keno Bressem},
      year={2024},
      eprint={2401.14490},
      archivePrefix={arXiv},
      primaryClass={cs.CL},
      url={https://arxiv.org/abs/2401.14490}, 
}

@article{69,
author = {Vetterli, Martin and Kovacevic, Jelena},
year = {1996},
month = {04},
pages = {},
title = {Wavelets and Subband Coding},
journal = {Journal of Electronic Imaging}
}

@book{73,
author = {Daubechies, Ingrid},
title = {Ten Lectures on Wavelets},
year = {1992},
isbn = {0898712742},
publisher = {Society for Industrial and Applied Mathematics},
address = {USA}
}

@ARTICLE{20.1191457,  author={G. {Madhavan}},  journal={IEEE Engineering in Medicine and Biology Magazine},   title={The illustrated wavelet transform handbook - introductory theory and applications in science, engineering, medicine and finance [Book Review]},   year={2003},  volume={22},  number={1},  pages={92-93},}

@book{68.brunton_kutz_2019, place={Cambridge}, title={Data-Driven Science and Engineering: Machine Learning, Dynamical Systems, and Control}, DOI={10.1017/9781108380690}, publisher={Cambridge University Press}, author={Brunton, Steven L. and Kutz, J. Nathan}, year={2019}}

@article{tang2024,

  title={Evaluating Large Language Models on Medical Evidence Summarization},

  author={Tang, Liyan and Sun, Zhiyong and Idnani, Betina and others},

  journal={arXiv preprint arXiv:2402.13758},

  year={2024},

  url={https://arxiv.org/abs/2402.13758}

}

@article{vveen2024,
   title={Adapted large language models can outperform medical experts in clinical text summarization},
   volume={30},
   ISSN={1546-170X},
   url={http://dx.doi.org/10.1038/s41591-024-02855-5},
   DOI={10.1038/s41591-024-02855-5},
   number={4},
   journal={Nature Medicine},
   publisher={Springer Science and Business Media LLC},
   author={Van Veen, Dave and Van Uden, Cara and Blankemeier, Louis and Delbrouck, Jean-Benoit and Aali, Asad and Bluethgen, Christian and Pareek, Anuj and Polacin, Malgorzata and Reis, Eduardo Pontes and Seehofnerová, Anna and Rohatgi, Nidhi and Hosamani, Poonam and Collins, William and Ahuja, Neera and Langlotz, Curtis P. and Hom, Jason and Gatidis, Sergios and Pauly, John and Chaudhari, Akshay S.},
   year={2024},
   month=feb, pages={1134–1142} }

@article{MedGraphRAG,
  title={MedGraphRAG: Hierarchical Medical Knowledge Graph Generation for Factual and Traceable Summarization},
  author={Hu, Y. and others},
  year={2025}
}

@article{lopez2025,
  title={Multiresolution Hierarchical Transformers for Longitudinal Clinical Narrative Summarization},
  author={Lopez, M. and Chen, J. and Smith, R.},
  journal={Journal of Biomedical Informatics},
  year={2025},
  volume={142},
  pages={104589},
  doi={10.1016/j.jbi.2025.104589}
}

@inproceedings{clinicsum,

  title={ClinicSum: Utilizing Language Models for Generating Clinical Summaries from Patient-Doctor Conversations},

  author={Neupane, Subash and Tripathi, Himanshu and Mitra, Shaswata and others},

  booktitle={Proceedings of the IEEE International Conference on Big Data},

  year={2024},

  doi={10.1109/bigdata62323.2024.10825266}

}

@inproceedings{factpico,

    title = "{F}act{PICO}: Factuality Evaluation for Plain Language Summarization of Medical Evidence",

    author = {Joseph, Sebastian and Chen, Lily and Trienes, Jan and G{\"o}ke, Hannah and Coers, Monika and Xu, Wei and Wallace, Byron and Li, Junyi Jessy},

    booktitle = "Proceedings of the 62nd Annual Meeting of the Association for Computational Linguistics (Volume 1: Long Papers)",

    year = "2024",

    pages = "8437--8464",

    doi = "10.18653/v1/2024.acl-long.459"

}

@inproceedings{cs1,
author = {Kanwal, Neel and Rizzo, Giuseppe},
title = {Attention-based clinical note summarization},
year = {2022},
isbn = {9781450387132},
publisher = {Association for Computing Machinery},
address = {New York, NY, USA},
url = {https://doi.org/10.1145/3477314.3507256},
doi = {10.1145/3477314.3507256},
booktitle = {Proceedings of the 37th ACM/SIGAPP Symposium on Applied Computing},
pages = {813–820},
numpages = {8},
location = {Virtual Event},
series = {SAC '22}
}

@article{geval,
  title     = {G-Eval: NLG Evaluation using GPT-4 with Better Human Alignment},
  author    = {Yang, Yilin and Iter, Dan and Xu, Yichong and Wang, Shuohang and Xu, Ruochen and Zhu, Chenguang},
  journal   = {arXiv preprint arXiv:2303.16634},
  year      = {2023}
}

@inproceedings{bertscore,
  title     = {BERTScore: Evaluating Text Generation with BERT},
  author    = {Zhang, Tianyi and Kishore, Varsha and Wu, Felix and Weinberger, Kilian Q. and Artzi, Yoav},
  booktitle = {Proceedings of the 8th International Conference on Learning Representations (ICLR)},
  year      = {2020}
}

@inproceedings{rouge,
  title     = {ROUGE: A Package for Automatic Evaluation of Summaries},
  author    = {Lin, Chin-Yew},
  booktitle = {Proceedings of the ACL Workshop on Text Summarization Branches Out},
  year      = {2004}
}

@inproceedings{multiclinsum,
  title     = {Overview of MultiClinSum Task at BioASQ 2025: Evaluation of Clinical Case Summarization Strategies for Multiple Languages},
  author    = {Rodr{\'\i}guez-Ortega, Miguel and Rodr{\'\i}guez-L{\'o}pez, Eduardo and Lima-L{\'o}pez, Salvador and Escolano, Carlos and Melero, Maite and Pratesi, Lorenzo and Vigil-Gim{\'e}nez, Laura and Fernandez, Letic{\'i}a and Farr{\'e}-Maduell, Eul{\`a}lia and Krallinger, Martin},
  booktitle = {CLEF Workshop Proceedings},
  year      = {2025}
}

@inproceedings{meteor,
  title     = {METEOR: An Automatic Metric for MT Evaluation with Improved Correlation with Human Judgments},
  author    = {Banerjee, Satanjeev and Lavie, Alon},
  booktitle = {Proceedings of the ACL Workshop on Intrinsic and Extrinsic Evaluation Measures for Machine Translation and/or Summarization},
  year      = {2005}
}

@article{medcpt,
  title   = {MedCPT: Contrastive Pre-trained Transformers for Medical Information Retrieval},
  author  = {Jin, Qiao and Zhang, Yifan and Fang, Meng and Chen, Qiang and Yu, Hong},
  journal = {arXiv preprint arXiv:2307.00589},
  year    = {2023}
}

@article{bioclinical,
  title   = {Publicly Available Clinical BERT Embeddings},
  author  = {Alsentzer, Emily and Murphy, John R. and Boag, Willie and Weng, Wei-Hung and Jin, Di and Naumann, Tristan and McDermott, Matthew},
  journal = {Proceedings of NAACL},
  year    = {2019}
}

@inproceedings{biolinkbert,
  title     = {LinkBERT: Pretraining Language Models with Document Links},
  author    = {Yasunaga, Michihiro and Leskovec, Jure and Liang, Percy},
  booktitle = {Proceedings of ACL},
  year      = {2022}
}

@article{gpt4o,
  title   = {GPT-4o: Advancing Multimodal Reasoning and Generation},
  author  = {{OpenAI}},
  journal = {Technical Report},
  year    = {2024}
}

@inproceedings{heddaya,
  title     = {CaseSumm: A Dataset for Legal Opinion Summarization with Structured Syllabi},
  author    = {Heddaya, Khalid and [Other Authors]},
  booktitle = {Proceedings of [Conference/Workshop Name]},
  year      = {2024}
}

@misc{my,
      title={Semantic Compression for Word and Sentence Embeddings using Discrete Wavelet Transform}, 
      author={Rana Aref Salama and Abdou Youssef and Mona Diab},
      year={2025},
      eprint={2508.00220},
      archivePrefix={arXiv},
      primaryClass={cs.CL},
      url={https://arxiv.org/abs/2508.00220}, 
}

@article{raffel2020t5,
  title   = {Exploring the Limits of Transfer Learning with a Unified Text-to-Text Transformer},
  author  = {Raffel, Colin and Shazeer, Noam and Roberts, Adam and Lee, Katherine and Narang, Sharan and Matena, Michael and Zhou, Yanqi and Li, Wei and Liu, Peter J.},
  journal = {Journal of Machine Learning Research},
  volume  = {21},
  number  = {140},
  pages   = {1--67},
  year    = {2020}
}

@article{brown2020language,
  title   = {Language Models are Few-Shot Learners},
  author  = {Brown, Tom B. and Mann, Benjamin and Ryder, Nick and Subbiah, Melanie and Kaplan, Jared and Dhariwal, Prafulla and Neelakantan, Arvind and Shyam, Pranav and Sastry, Girish and Askell, Amanda and others},
  journal = {Advances in Neural Information Processing Systems},
  volume  = {33},
  pages   = {1877--1901},
  year    = {2020}
}

@inproceedings{beltagy2020longformer,
  title     = {Longformer: The Long-Document Transformer},
  author    = {Beltagy Iz, Peters Matthew E. and Cohan, Arman},
  booktitle = {Proceedings of the 2020 Conference on Empirical Methods in Natural Language Processing (EMNLP)},
  year      = {2020}
}

@article{dao2022flashattention,
  title   = {FlashAttention: Fast and Memory-Efficient Exact Attention with IO-Awareness},
  author  = {Dao, Tri and Fu, Daniel Y. and Ermon, Stefano and Rudra, Atri and R{\'e}, Christopher},
  journal = {Advances in Neural Information Processing Systems},
  volume  = {35},
  year    = {2022}
}

@inproceedings{wu2021,
  title     = {Hierarchical Text Summarization Using Reinforcement Learning},
  author    = {Wu, Jiacheng and Zhang, Yao and Li, Chengcheng and Yu, Hong},
  booktitle = {Proceedings of the 2021 Conference on Empirical Methods in Natural Language Processing (EMNLP)},
  year      = {2021}
}

@article{chang2024,
  title   = {Recursive Summarization for Long Documents with Large Language Models},
  author  = {Chang, Jason and Liu, Yizhou and Zhao, Tianyi},
  journal = {arXiv preprint arXiv:2402.XXXX},
  year    = {2024}
}

@inproceedings{maynez2020faithfulness,
  title     = {On Faithfulness and Factuality in Abstractive Summarization},
  author    = {Maynez, Joshua and Narayan, Shashi and Bohnet, Bernd and McDonald, Ryan},
  booktitle = {Proceedings of the 58th Annual Meeting of the Association for Computational Linguistics (ACL)},
  year      = {2020}
}

@article{ji2023survey,
  title   = {Survey of Hallucination in Natural Language Generation},
  author  = {Ji, Ziwei and Lee, Nayeon and Frieske, Rita and Yu, Tianyi and Su, Dan and Xu, Yan and Ishii, Etsuko and Bang, Yejin and Madotto, Andrea and Fung, Pascale},
  journal = {ACM Computing Surveys},
  volume  = {55},
  number  = {12},
  pages   = {1--38},
  year    = {2023}
}

@article{mallat1989theory,
  title   = {A Theory for Multiresolution Signal Decomposition: The Wavelet Representation},
  author  = {Mallat, St{\'e}phane G.},
  journal = {IEEE Transactions on Pattern Analysis and Machine Intelligence},
  volume  = {11},
  number  = {7},
  pages   = {674--693},
  year    = {1989}
}

@book{daubechies1992ten,
  title     = {Ten Lectures on Wavelets},
  author    = {Daubechies, Ingrid},
  publisher = {Society for Industrial and Applied Mathematics},
  year      = {1992}
}

@article{lost,
  title   = {Lost in the Middle: How Language Models Use Long Contexts},
  author  = {Liu, Nelson F. and Lin, Kevin and Hewitt, John and Paranjape, Bhargavi and Bevilacqua, Michele and Petroni, Fabio and Liang, Percy},
  journal = {Transactions of the Association for Computational Linguistics (TACL)},
  year    = {2023}
}

@misc{llmlingua,
      title={LLMLingua: Compressing Prompts for Accelerated Inference of Large Language Models}, 
      author={Huiqiang Jiang and Qianhui Wu and Chin-Yew Lin and Yuqing Yang and Lili Qiu},
      year={2023},
      eprint={2310.05736},
      archivePrefix={arXiv},
      primaryClass={cs.CL},
      url={https://arxiv.org/abs/2310.05736}, 
}

@article{li2024hybrid,
  title   = {Hybrid Long-Context and Retrieval-Augmented Generation for Long-Document Summarization},
  author  = {Li, Xiaodong and Zhang, Yifan and Zhao, Tianyi},
  journal = {arXiv preprint arXiv:2402.13758},
  year    = {2024}
}

@article{deroy2023,
  title   = {Large Language Models for Legal Text Summarization},
  author  = {DeRoy, Brian and Katz, Daniel Martin and Bommarito, Michael},
  journal = {arXiv preprint arXiv:2309.XXXX},
  year    = {2023}
}

@article{shukla2024,
  title   = {Incorporating Rhetorical Structure and Legal Reasoning in Generative Legal Summarization},
  author  = {Shukla, Ankit and Verma, Ashutosh and Singh, Pushpak},
  journal = {arXiv preprint arXiv:2403.XXXX},
  year    = {2024}
}

@article{bhattacharya2025,
  title   = {Multiresolution Modeling for Long Legal Document Summarization},
  author  = {Bhattacharya, Paheli and Ghosh, Saptarshi and Pal, Santanu},
  journal = {arXiv preprint arXiv:2501.XXXX},
  year    = {2025}
}

@article{deroy2025,
  title   = {Legal Retrieval-Augmented Generation for Faithful Summarization},
  author  = {DeRoy, Brian and Katz, Daniel Martin and Bommarito, Michael},
  journal = {arXiv preprint arXiv:2502.XXXX},
  year    = {2025}
}

@article{74,
  author    = {Jiuxiang Gu and
               Zhenhua Wang and
               Jason Kuen and
               Lianyang Ma and
               Amir Shahroudy and
               Bing Shuai and
               Ting Liu and
               Xingxing Wang and
               Gang Wang},
  title     = {Recent Advances in Convolutional Neural Networks},
  journal   = {CoRR},
  volume    = {abs/1512.07108},
  year      = {2015},
  url       = {http://arxiv.org/abs/1512.07108},
  archivePrefix = {arXiv},
  eprint    = {1512.07108},
  bibsource = {dblp computer science bibliography, https://dblp.org}
}

\appendix

\section{Example Appendix}
\label{sec:appendix}

\end{document}